\documentclass[letterpaper]{article} 
\usepackage{aaai25}  
\usepackage{times}  
\usepackage{helvet}  
\usepackage{courier}  
\usepackage[hyphens]{url}  
\usepackage{graphicx} 
\usepackage{natbib}  
\usepackage{caption} 
\usepackage{algorithm}
\usepackage{algorithmic}
\usepackage{newfloat}
\usepackage{listings}
\DeclareCaptionStyle{ruled}{labelfont=normalfont,labelsep=colon,strut=off} 
\floatstyle{ruled}
\newfloat{listing}{tb}{lst}{}
\floatname{listing}{Listing}
\usepackage{cite}
\usepackage{amsmath,amssymb,amsfonts}
\usepackage{algorithmic}
\usepackage{graphicx}
\usepackage{textcomp}
\usepackage{xcolor}
\usepackage{color}

\definecolor{mycolorblack}{RGB}{84,89,105}    
\definecolor{mycolorbrown}{RGB}{164,117,125}  
\definecolor{mycolororange}{RGB}{231,152,124} 
\definecolor{mycolorblue}{RGB}{139,145,182}   
\definecolor{mycolorpurple}{RGB}{119,113,164}

\definecolor{mycolor41}{RGB}{54,80,131}
\definecolor{mycolor42}{RGB}{183,131,175}
\definecolor{mycolor43}{RGB}{245,166,115}
\definecolor{mycolor44}{RGB}{252,219,114}

\definecolor{mycolor1}{RGB}{17,50,93}
\definecolor{mycolor2}{RGB}{54,80,131}
\definecolor{mycolor3}{RGB}{115,107,157}
\definecolor{mycolor4}{RGB}{183,131,175}
\definecolor{mycolor5}{RGB}{245,166,115}
\definecolor{mycolor6}{RGB}{252,219,114}

\definecolor{mycolor91}{RGB}{54,80,131}
\definecolor{mycolor92}{RGB}{128, 255, 128}
\definecolor{mycolor93}{RGB}{128, 128, 255}
\definecolor{mycolor94}{RGB}{17,50,93}
\definecolor{mycolor95}{RGB}{192, 128, 192}
\definecolor{mycolor96}{RGB}{128, 255, 255}
\definecolor{mycolor97}{RGB}{255, 207, 128}
\definecolor{mycolor98}{RGB}{255, 180, 215}
\definecolor{mycolor99}{RGB}{197, 134, 109}

\usepackage{csvsimple}

\makeatother 

\usepackage{tikz}
\usepackage{pgfplots}
\pgfplotsset{width=8cm,compat=1.13} 
\usepackage{arydshln} 

\usepackage{subfigure}

\usepackage{enumitem}

\newcommand{\name}[0]{{GARLIC}}

\newcommand{\revise}[1]{{\color{black} #1}}

\newcommand{\etal}{{\emph{et al.}\ }}
\newcommand{\eg}{{\emph{e.g.,}\ }}
\newcommand{\ie}{{\emph{i.e.,}\ }}

\title{GARLIC: GPT-Augmented Reinforcement Learning\\ with Intelligent Control for Vehicle Dispatching}

\author{
    Xiao Han\textsuperscript{\rm 1},
    Zijian Zhang\textsuperscript{\rm 2},
    Xiangyu Zhao\textsuperscript{\rm 1},
    Yuanshao Zhu\textsuperscript{\rm 1},
    Guojiang Shen\textsuperscript{\rm 3},\\
    Xiangjie Kong\textsuperscript{\rm 3},
    Xuetao Wei\textsuperscript{\rm 4},
    Liqiang Nie\textsuperscript{\rm 5},
    Jieping Ye\textsuperscript{\rm 6}
}
\affiliations{
    \textsuperscript{\rm 1}City University of Hong Kong\quad
    \textsuperscript{\rm 2}Jilin University\quad
    \textsuperscript{\rm 3}Zhejiang University of Technology\\
    \textsuperscript{\rm 4}Southern University of Science and Technology\quad
    \textsuperscript{\rm 5}Harbin Institute of Technology (Shenzhen)\quad
    \textsuperscript{\rm 6}Alibaba Group\\
    hahahenha@gmail.com, zhangzijian@jlu.edu.cn, xianzhao@cityu.edu.hk, yaso.zhu@my.cityu.edu.hk, gjshen1975@zjut.edu.cn, xjkong@ieee.org, weixt@sustech.edu.cn, nieliqiang@gmail.com, jieping@gmail.com

}

\begin{document}

\maketitle

\begin{abstract}

As urban residents demand higher travel quality, vehicle dispatch has become a critical component of online ride-hailing services. However, current vehicle dispatch systems struggle to navigate the complexities of urban traffic dynamics, including unpredictable traffic conditions, diverse driver behaviors, and fluctuating supply and demand patterns. These challenges have resulted in travel difficulties for passengers in certain areas, while many drivers in other areas are unable to secure orders, leading to a decline in the overall quality of urban transportation services.
To address these issues, this paper introduces \name: a framework of GPT-Augmented Reinforcement Learning with Intelligent Control for vehicle dispatching. \name\ utilizes multiview graphs to capture hierarchical traffic states, and learns 
a dynamic reward function that accounts for individual driving behaviors.
The framework further integrates a GPT model trained with a custom loss function to enable high-precision predictions and optimize dispatching policies in real-world scenarios. Experiments conducted on two real-world datasets demonstrate that \name\ effectively aligns with driver behaviors while reducing the empty load rate of vehicles.

\end{abstract}

\section{Introduction}

The past decade has witnessed explosive growth in online car-hailing services, fundamentally transforming urban transportation.
Central to this transformation is the role of vehicle dispatching \cite{shi2024vehicle}, which serves as a pivotal component in reducing the waiting time of passengers, increasing the income of drivers, and facilitating daily transportation \cite{barrios2023cost,rahman2023impacts,sadrani2022vehicle}.
In recent years, reinforcement learning (RL) methods have emerged as outstanding performers in areas such as multi-agent control and sequential decision-making \cite{qiu2023hierarchical,ellis2024smacv2,han2023mitigating}.
Therefore, many studies have leveraged RL techniques to enhance vehicle dispatching, treating it as a multi-agent sequential decision-making task \cite{guo2024enhancing,huang2023effective}.

\begin{figure}[t]
\centering
\subfigure[Multi-hop communication for environmental perception]{
\includegraphics[width=0.45\linewidth]{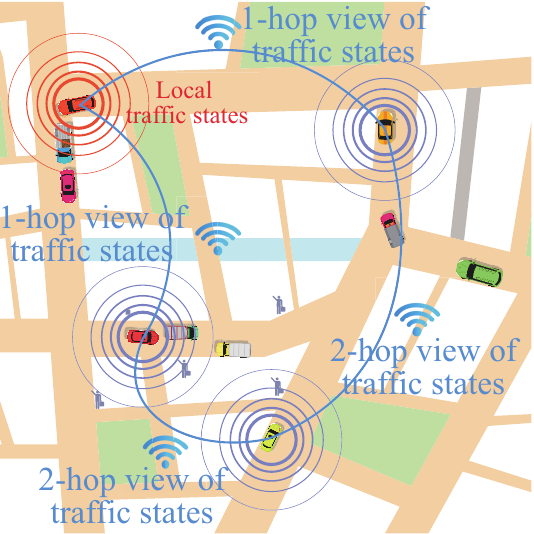}
\label{fig:1a}
}
\subfigure[Latency of multi-hop 5G V2V communication]{
\includegraphics[width=0.45\linewidth]{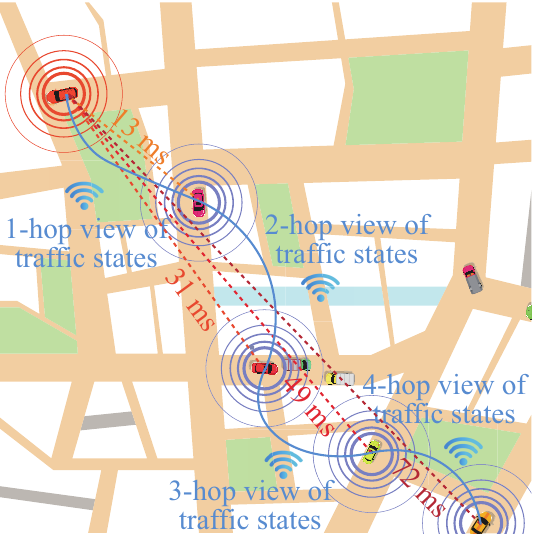}
\label{fig:1b}
}

\subfigure[A case of vehicle dispatching]{
\includegraphics[width=0.55\linewidth]{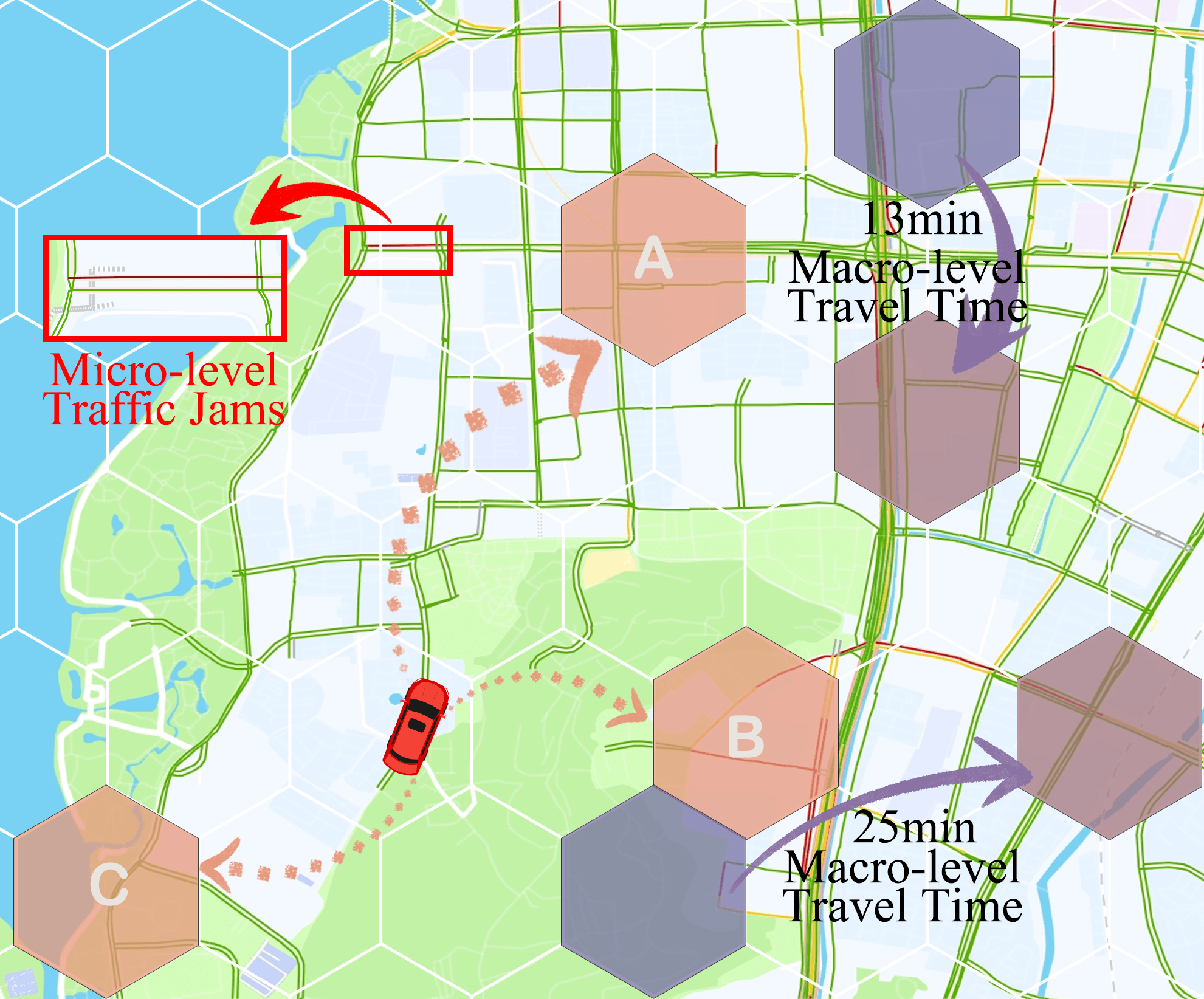}
\label{fig:1c}
}
\caption{A vehicle dispatching scenario.}
\label{fig:1}
\vspace{-6mm}
\end{figure}

However, unlike traditional multi-agent reinforcement learning (MARL) approaches applied in other domains, vehicle dispatching presents a unique challenge due to the complex interplay between observable local traffic states and undetectable global spatiotemporal correlations.
Each vehicle acts as an individual agent, with access limited to the environmental states in its immediate vicinity.
This makes it difficult to obtain a comprehensive, global view of vehicle supply and demand.
As illustrated in Figure \ref{fig:1a}, a vehicle must rely on multiple hops of vehicle-to-vehicle (V2V) communication to acquire more extensive traffic flow information.
Furthermore, expanding a vehicle's receptive field exponentially increases communication latency among agents \cite{huang2022k,wang2023optimizing}, as depicted in Figure \ref{fig:1b}.
According to the traffic flow theory \cite{gerlough1976traffic}, traffic flows also behave differently at diverse granularities.
For instance, macro-level traffic flow provides an overview of travel times, as shown by the arrows between purple and brown grids in Figure \ref{fig:1c}.
In contrast, micro-level traffic states can pinpoint traffic jams directly, as illustrated by the different road segment colors (green, yellow, and red) in Figure \ref{fig:1c}.
In summary, obtaining a comprehensive and accurate view of traffic states is a significant challenge in vehicle dispatching.

Accurate vehicle dispatching also necessitates nuanced driving behavior modeling, which accounts for the individual preferences of different vehicle agents regarding dispatching instructions. 
Driving behavior reflects the
driver’s personal inclination toward specific dispatching tasks, and plays a crucial yet often overlooked role in transportation \cite{wang2024reinforcement,robbennolt2023maximum,zhang2023autostl,DBLP:conf/kdd/HanZZW23}.
For example, consider the taxi driver of the red car in Figure \ref{fig:1c}, who is more familiar with region A.
This driver might prefer to pick up passengers in region A rather than in the unfamiliar regions B or C, even if those regions are closer.
Consequently, a dispatching algorithm that ignores drivers' behavior patterns may disrupt the overall traffic system.

To address the all above challenges, we propose a \underline{\textbf{G}}PT-\underline{\textbf{A}}ugmented \underline{\textbf{R}}einforcement \underline{\textbf{L}}earning
with \underline{\textbf{I}}ntelligent \underline{\textbf{C}}ontrol framework, \textbf{\name}, which utilizes an improved MARL approach.
Specifically, we design a hierarchical traffic state representation module to integrate traffic features at different granularities, providing a comprehensive representation of real-time traffic conditions.
Additionally, we quantify driving behavior through dynamic rewards using a contrastive learning method, aligning dispatching instructions with the intents of drivers.
Given the complex analytical and understanding capabilities required for learning vehicle dispatching policies, we employ a Generative Pre-trained Transformer (GPT)-augmented model with a self-defined loss function to enhance the expression of the framework.
To the best of our knowledge, our innovative framework offers a comprehensive solution to the core challenges in vehicle dispatching, setting a new benchmark in this field.
Our main contributions can be summarized as follows:

\begin{itemize}

\item
Our proposed framework, \name, combines hierarchical traffic state representation, dynamic reward generation, and GPT-augmented dispatching policy learning.
To the best of our knowledge, this novel approach builds a complete GPT-enhanced MARL vehicle dispatching framework that has not been explored previously;

\item
We utilize multiview graphs to depict the hierarchical traffic states in the road networks and establish a dynamic reward model for capturing driving behaviors, leading to better dispatching policy outcomes. These innovations contribute significantly to the improved performance of vehicle dispatching;

\item
Extensive experiments on two real-world
road networks against advancing baselines demonstrate the effectiveness and efficiency of \name.

\end{itemize}

\section{Related Work}

This section provides a concise overview of related research in vehicle dispatching. Unlike car-hailing order dispatching, vehicle dispatching focuses on relocating vehicles to ensure a future balance between supply and demand. Many previous studies have modeled this as a Markov decision process, which relies on explicitly fitted state transition probabilities \cite{zhang2024automated,zhang2023vehicle,sun2024dynamic}.
To efficiently model this Markov decision process, RL has been widely applied to vehicle repositioning tasks \cite{DBLP:conf/aaai/ChenSSW0ZGL24,qin2022reinforcement}, where the global traffic state and reward function are used to enhance the precision of repositioning. However, the global traffic-state perception in existing methods has high communication latency, hindering real-time dispatch \cite{DBLP:journals/twc/ShiJXCZCW24}. To address this, we designed a multiview graph learning module with limited hops.


Furthermore, driver behavior plays a crucial role in transportation analysis \cite{cui2024survey,zhang2023mlpst,ma2023rethinking}. Recent studies have begun to incorporate driving behavior into driving applications. For example, Li \etal \cite{li2022driver} used IL method to replicate human driving behavior, effectively transferring these strategies to autonomous vehicle scenarios.
Jackson \etal \cite{jackson2024natural} uses the powerful analysis and processing capabilities of LLAMA-7B to characterize driving behavior and then uses it for autonomous driving simulation.
However, there is a relative scarcity of research that quantifies vehicle driving behavior to directly evaluate the rationality of vehicle dispatching orders. Consequently, there is an urgent need to design a more efficient and accurate driving behavior-based vehicle dispatching system.

\section{Preliminary}

In this paper, we adopt a novel MARL method to optimize online car-hailing dispatching policies.
This section outlines critical definitions for understanding our paper.

\noindent \textbf{Vehicle Trajectory} $\tau$:
This refers to a sequence of GPS points $(x_t, y_t)$ recorded over a time interval $t \in [T]$, represented as $\tau = { (x_1,y_1,t_1), \cdots, (x_T, y_T, t_T) }$. A vehicle can generate multiple trajectories based on different statuses (such as empty or occupied).
We focus solely on empty vehicle trajectories to better understand driving behavior when drivers don't have a specific destination.

\noindent \textbf{multiview Graph} $\mathcal{G}^i$:
We define the multiview graph as $\mathcal{G}^i = \{ \boldsymbol{V}^i, \boldsymbol{E}^i \}$, where $i \in \{ \text{micro,meso,macro} \}$ presents different views, the node set $\boldsymbol{V}^i$ represents various traffic zones, and the edge set $\boldsymbol{E}^i$ indicates the connections among these zones. 
The features of each traffic zone at time $t$ are denoted by $\boldsymbol{X}^i_t \in \mathbb{R}^{|\boldsymbol{V}| \times m^i}$, capturing vehicle availability and order demand. For different views of graphs, we have different graph features: $m^{\text{micro}} \neq m^{\text{meso}} \neq m^{\text{macro}}$.

\begin{figure*}[htp!]
    \centering
    \includegraphics[width=0.85\linewidth]{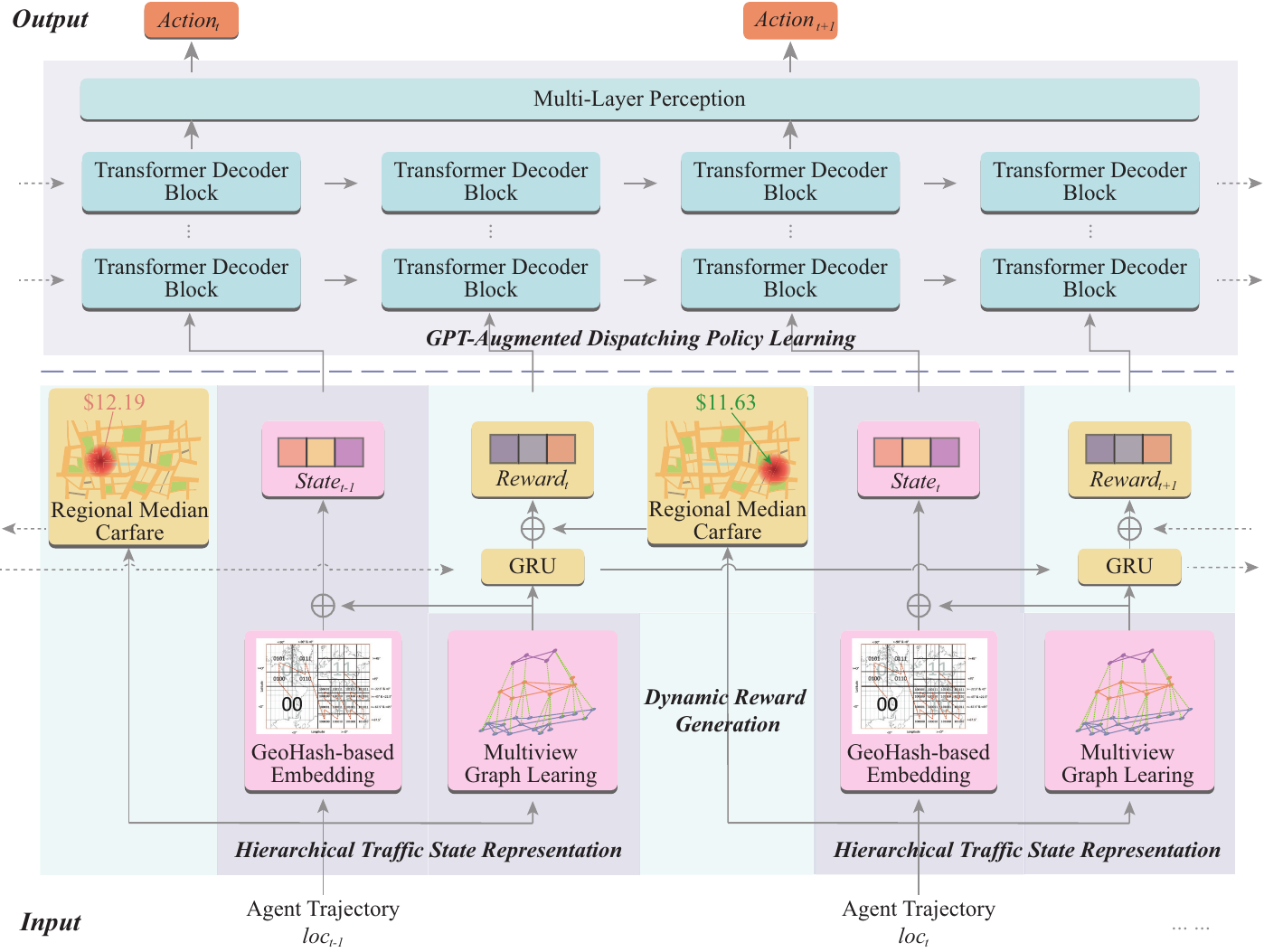}
    \caption{The framework overview of \name.
    }
    \label{fig:2}
\vspace{-6mm}
\end{figure*}

\noindent \textbf{Multi-Agent Reinforcement Learning for Vehicle Dispatching}:
In our model, each vehicle in the road network acts as an independent agent, with distinct driving behaviors and the ability to generate continuous trajectories and monitor local traffic conditions. For each agent (vehicle) $u$, we consider the following five essential elements:
\begin{itemize}[leftmargin=*]
    \item \textbf{Decision Time $[T]$}:
    It is a set of all finite decision timesteps $[T] = \{ 1, \cdots, t, \cdots, T\}$. At each timestep $t$, the vehicle location and environment states are sampled.

    \item \textbf{Action $\mathcal{A}^{u}$}:
    $\mathcal{A}^{u} = \{ a_0^{u}, \cdots, a_t^{u}, \cdots, a_T^{u}\}$ represent the set of actions to balance the vehicle supply and demand.
    $a_t^{u} := \{dis, deg\}$ is an action performed by the vehicle $u$ at time $t$, where
    $dis$ is the straight-line distance a vehicle needs to travel from time $t$ to $t+1$, and $deg$ means the azimuth angle between the target and current locations. 

    \item \textbf{State $\mathcal{S}^{u}$}:
    $\mathcal{S}^{u} = \{ \boldsymbol{S}_0^{u}, \cdots,  \boldsymbol{S}_t^{u}, \cdots,  \boldsymbol{S}_T^{u}\}$ represents the set of traffic states observed at each time $t$. Here $\boldsymbol{S}_t^{u}$ is the concatenation of the state embedding matrix $\boldsymbol{Emb}_{\mathcal{G},t}$ extracted from the traffic environment and the location embedding matrix $\boldsymbol{Emb}_{loc,t}^u$ of the vehicle at time $t$.

    \item \textbf{Reward $\mathcal{R}^{u}$}:
    $\mathcal{R}^{u} = \{ r_0^{u}, \cdots, r_t^{u}, \cdots, r_T^{u}\}$ represents the set of rewards calculated by the reward function, and it is predefined according to the driver's driving behavior and the taxi fare.
    The total return is defined as $\sum_t \gamma \cdot r_{t}^{u}$, where $\gamma$ is a discount factor, $\gamma \in [0, 1]$.

    \item \textbf{Policy $\pi_{\theta}^{u}$}:
    $\pi_{\theta}^{u} = \pi_\theta^{u} (a | s)$ is a mapping from traffic states to dispatching actions of the $u$-th agent.
    The policy $\pi$ determines the appropriate vehicle dispatch instructions $a$ by analyzing the state $s$, which includes various features of the environment and the current status of the agent.
\end{itemize}

While agents in the same area and close to each other may share the same multiview graphs $\mathcal{G}^i$ of the road network and observe similar traffic features $\boldsymbol{X}_t$, they exhibit unique driving behaviors that significantly influence their vehicle trajectories.
To account for these behavioral differences, our study departs from conventional MARL frameworks with fixed rewards by employing a dynamic reward model. Additionally, we propose a GPT-augmented MARL model to learn more effective dispatching policies.

\section{The Proposed Framework}

In this section, we first provide a framework overview of \name.
Then we introduce the hierarchical traffic state representation method to capture the real-time traffic states.
Furthermore, we demonstrate a dynamic driving reward generation approach to score vehicle trajectories under different driving behaviors.
Finally, a GPT-augmented dispatching policy learning model is applied to combine all of the components and learn the vehicle dispatching policy.

\subsection{Overview}

Figure \ref{fig:2} provides an illustration of the overall vehicle dispatching framework, which is composed of three key modules: 
the hierarchical traffic state representation module,
the dynamic reward generation module,
and the GPT-augmented dispatching policy learning module.

In the first module, we employ a multiview Graph Convolutional Network (GCN) to represent the hierarchical traffic state by integrating traffic information gathered by various vehicles at different levels of granularity. By combining this with GeoHash-based vehicle location embeddings, we can accurately calculate the real-time traffic state of the specific region where each vehicle is located.

The second module utilizes a  Gated Recurrent Unit (GRU)-based Recurrent Neural Network (RNN) to model driving behaviors, generating dynamic rewards that are weighted by the regional median carfare. This approach ensures that the reward system reflects both the temporal and spatial nuances of driver behavior.

Finally, in the third module, we frame the training of the MARL-based vehicle dispatching task as a supervised learning process \cite{wang2024critic,yamagata2023q}.
For each agent, the time-ordered states and rewards are utilized as inputs, and a GPT-augmented model is employed to produce high-precision actions for vehicle dispatching.

\subsection{Hierarchical Traffic State Representation}

Urban spatiotemporal data exhibits hierarchical characteristics \cite{ning2024uukg,zhang2023promptst,han2020congestion}, which cannot be directly represented using a single structured data format.
For instance, features such as turning movements at a crossroad can only be captured from a micro-level view of the traffic environment, whereas the average travel time is a feature observable only from a macro-level perspective of the same environment.
These features differ in sampling frequencies, dimensions, and units, necessitating specialized approaches to represent and integrate them accurately.

\begin{figure}[htp!]
    \centering
    \subfigure[The multi-level road network modeling]{
    \label{fig:3a}
    \includegraphics[width=0.45\linewidth]{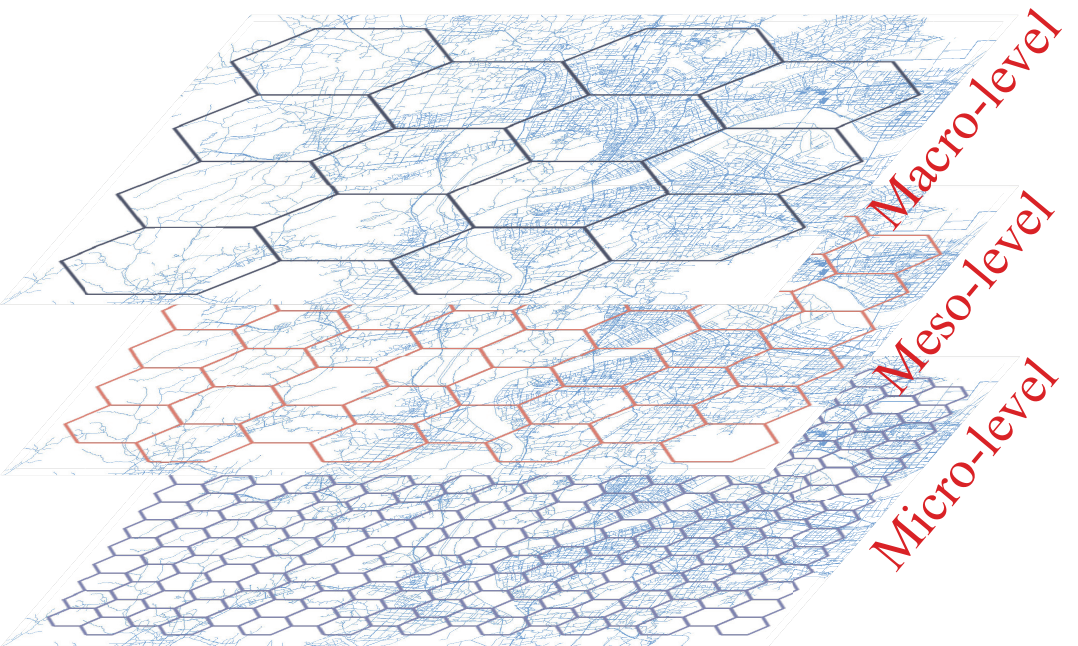}
    }
    \subfigure[The multiview graph construction]{
    \label{fig:3b}
    \includegraphics[width=0.45\linewidth]{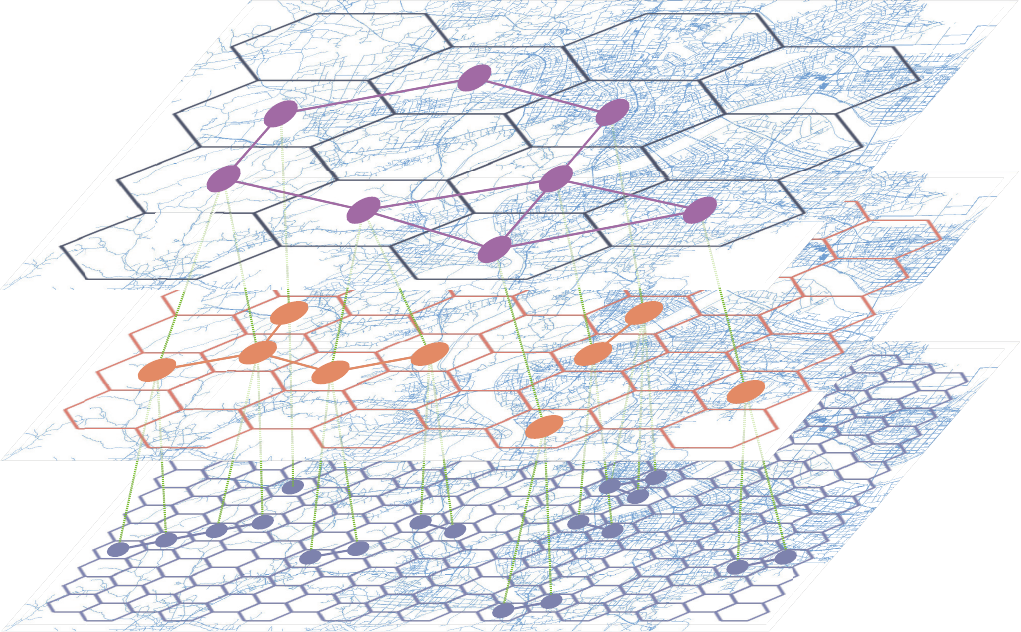}
    }
    \caption{The multiview graph of road networks.}
    \label{fig:3}
\end{figure}

To address this issue, 
we present the road network as multiview honeycomb graphs,
as shown in Figure \ref{fig:3a}.
In this representation, the road network is divided into grids comprising square hexagons of varying radii, each representing a distinct view.
Here, the hexagon-based grids ensure uniform distance from all adjacent neighbors to the central grid, facilitating more precise modeling of different spatial regions than square grid-based methods.
To construct the multiview graph, each grid is treated as a node, and the traffic information in a grid is considered to be the node feature, with edges connecting adjacent grids, as shown in Figure \ref{fig:3b}.

Unlike other road network modeling methods, we calculate distinct traffic indicators for different views of graphs.
The micro-level graph primarily utilizes vehicle trajectory, road congestion status, and vehicle speed—data that can be directly obtained from the local environment ($radii \leq 1 \text{km}$) of a vehicle. The meso-level graph considers factors such as traffic volume, average traffic speed, 
intersection performance, 
and parking availability, which require analysis of all vehicles passing through a set of certain traffic sections\footnote{The ``traffic section" refers to a specific segment of a road or highway between two points, often delineated by intersections, junctions, or other distinct markers.} ($1 \text{km} < radii < 5 \text{km}$).
Meanwhile, the macro-level graph includes features such as average travel time, 
road network connectivity, and overall traffic conditions, which necessitate a more comprehensive analysis of vehicles across a broader range ($radii \ge 5 \text{km}$) of the road network.


To extract refined the traffic embeddings $\boldsymbol{Emb}_{\mathcal{G}_t^u}^u$, a GCN-based model is then deployed for multiview graph representation for a vehicle $u$:

\begin{equation}
\label{eq:1}
\centering
\begin{aligned}
\boldsymbol{Emb}_{\mathcal{G}_t^u}^u =& \mathrm{Concat} \left( \boldsymbol{Emb}_{\mathcal{G}^{u,i}_t}^u \right), \\
\boldsymbol{Emb}_{\mathcal{G}^{u,i}_t}^u =& \mathrm{GCN}(\boldsymbol{A}^{u,i}, \boldsymbol{X}^{u,i}_t) = \boldsymbol{A}^{u,i}\boldsymbol{X}^{u,i}_t\boldsymbol{W}^{i}, 
\end{aligned}
\end{equation}
where $i \in \{ \text{micro},\text{meso},\text{macro} \}$, $\boldsymbol{W}^{i}$ is the weight matrix that need to be trained, $\boldsymbol{A}^{u,i}$ is the adjacency matrix of the graph $\mathcal{G}^{u,i}$ under a specific view $i$, and $\boldsymbol{X}^{u,i}_t$ is the traffic features related to the graph $\mathcal{G}^{u,i}$ at time $t$.

Note that a high-accuracy location embedding of real-time trajectories is essential for this task.
We first use GeoHash \cite{morton1966computer} to encode each real-time GPS point, based on latitude and longitude, in the trajectory:

\begin{equation}
\label{eq:2}
\centering
\boldsymbol{Emb}_{loc_t^u}^u = \mathrm{GeoHash}\left( lat_t^u, lon_t^u \right),
\end{equation}
where $t \in \{0, 1, \cdots, T\}$ presents a specific timestep, and $loc_t^u:=(lat_t^u, lon_t^u)$ is the real-time GPS point of vehicle $u$.

By combining this location embedding with the traffic state surrounding vehicle $u$ at time $t$, we obtain the overall state embeddings of vehicle $u$:

\begin{equation}
\label{eq:3}
s_t^u = \mathrm{Concat}(\boldsymbol{Emb}_{\mathcal{G}_t^u}^u, \boldsymbol{Emb}_{loc_t^u}^u).
\end{equation}

\subsection{Dynamic Reward Generation}

The effective implementation of vehicle dispatching in real-world scenarios is largely influenced by driving behavior. However, early studies often ignored the quantification of driving behavior, focusing instead on minimizing vehicle imbalance or maximizing benefits in dispatching optimization \cite{wagenmaker2023leveraging}.
In this section, we propose a dynamic reward generation method that incorporates both driving behaviors and anticipated income. It quantifies the likelihood of drivers adhering to their driving habits by analyzing the vehicle trajectories in real-time.

To accurately capture the relationship between driving trajectories and corresponding driving behaviors in traffic embeddings, we deploy a GRU-based RNN model.
This network calculates the probability $p_t^{u}$ whether a trajectory belongs to a given vehicle $u$.

\begin{equation}
\label{eq:4}
\centering
\begin{aligned}
z_t^u =& \sigma \left( \boldsymbol{W}_{zx} \boldsymbol{Emb}_{loc_t^u}^u + \boldsymbol{W}_{zp} h_{t-1}^u + \boldsymbol{b}_z \right), \\
y_t^u =& \sigma \left( \boldsymbol{W}_{yx} \boldsymbol{Emb}_{loc_t^u}^u + \boldsymbol{W}_{yp} h_{t-1}^u + \boldsymbol{b}_y \right), \\
{p_t^u}' =& \tanh \left( \boldsymbol{W}_x' \boldsymbol{Emb}_{loc_t^u}^u + y_t^u \odot \boldsymbol{W}_p' p_{t-1}^u + \boldsymbol{b}' \right), \\
p_t^u =& z_t^u \odot p_{t-1}^u + (1 - z_t^u) \odot {p_t^u}',
\end{aligned}
\end{equation}
where $t \in \{1,2,\cdots, T\}$, $h_0:= \boldsymbol{Emb}_{loc_0}^u$ is the location embedding at initial timestep ($t = 0$), $\boldsymbol{W}_{\text{GRU}} = \left\{ \boldsymbol{W}_{zx}, \boldsymbol{W}_{zp}, \boldsymbol{W}_{yx}, \boldsymbol{W}_{yp},  \boldsymbol{W}_x', \boldsymbol{W}_p' \right\}$ is the set of weight matrices of GRU, and $\boldsymbol{b}_{\text{GRU}} = \left\{ 
\boldsymbol{b}_z, \boldsymbol{b}_y, \boldsymbol{b}' \right\}$ is the set of bias.

We employ a contrastive learning method to optimize the model parameters. Since different drivers exhibit distinct driving behaviors, trajectories generated by other vehicles are used as negative samples when modeling a specific driver’s behavior, as illustrated in Equation~\eqref{eq:contrastive}. 

\begin{equation}
\label{eq:contrastive}
Loss_{\text{pre-training}} = \max \sum_{u \in [N]}{ \sum_{t \in [T]}{q_t^u \log {p_t^u}}},
\end{equation}
where $N$ is the total number of online car-hailing vehicles, $q_t^u \in \{ 0,1 \}$ is the ground truth of the GPS point generated by the vehicle $u$ at time $t$.


Additionally, when a vehicle is carrying passengers, the regional median carfare earned by a driver is another factor influencing vehicle dispatching.
Therefore, we introduce the dynamic reward function, which incorporates both factors by introducing a hyperparameter $\alpha$ to weigh them together.

\begin{equation}
\label{eq:5}
r_t^u = \alpha \cdot p_t^u + (1 - \alpha) \cdot \sigma(\boldsymbol{W}_{\text{fare}} \hat{x}_{\text{fare},t}),
\end{equation}
where $\sigma(\cdot)$ is the sigmoid activation function, $\hat{x}_{\text{fare},t} = \sum_{t'=t}^{T}{x_{\text{fare},t'}}$, and $\boldsymbol{W}_{\text{fare}}$ is the weight matrix.

\subsection{GPT-Augmented Dispatching Policy Learning}

As discussed in the framework overview, vehicle dispatching can be effectively modeled as a MARL problem, which can also be reformulated as a supervised learning task. In this context, states and rewards are treated as sequential input data, with the corresponding sequence of actions as the output data.
However, the complexity of transportation systems, which requires the analysis of intricate traffic states and driving behaviors, demands advanced reasoning capabilities.
Recently, the GPT model has demonstrated strong performance in handling long-sequence, context-dependent, and structured data. Therefore, we utilize a GPT-augmented model to address these challenges.

Note that the core part of a GPT model is the transformer structure.
The input of the transformer is a sequence of temporal data, and we assign different positional embeddings to the data at different timesteps.
At each timestep, we mainly use two deep transformer decoder blocks to extract the probability of the next action.
We use the expected reward at the current time step as the input of the first transformer decoder block, and get the output embedding to guide the subsequent transformer decoder block to calculate the result:

\begin{equation}
\label{eq:6}
\begin{aligned}
\boldsymbol{Emb}_{r_t} &= \mathrm{Decoder}_{\mathrm{T}}(\boldsymbol{P}_{a_{t-1}} ,r_t, t), \\
\boldsymbol{P}_{a_t} &= \mathrm{Decoder}_{\mathrm{T}}(\boldsymbol{Emb}_{r_t}, s_t, t),
\end{aligned}
\end{equation}
where $t$ starts from 1, and $\boldsymbol{P}_{a_0} = \mathbf{0}$ is initialized as the zero tensor at $t=0$. $\mathrm{Decoder}_{\mathrm{T}}(x,y,z) = \mathrm{Decoder}^{k}_{\mathrm{T}} \odot \mathrm{Decoder}^{k-1}_{\mathrm{T}} \odot \cdots \odot \mathrm{Decoder}^{(1)}_{\mathrm{T}}(x,y,z)$ is a $k$-layer deep neural network of the transformer decoders. For each layer
$\mathrm{Decoder}^{(l)}_{\mathrm{T}}(x,y,t) = \mathrm{Attention}(x+\boldsymbol{Emb}_{pos}(t),x+\boldsymbol{Emb}_{pos}(t),y+\boldsymbol{Emb}_{pos}(t))$, we add the same positional embedding $\boldsymbol{Emb}_{pos}(t)$ to each input $x$ and $y$. Here $\mathrm{Attention}(x,y,z) = \mathrm{softmax}{(\frac{\sigma(x\boldsymbol{W}_x) \sigma(y \boldsymbol{W}_y)^{\top}}{\sqrt{d_y}})} \sigma(z \boldsymbol{W}_z)$, where $\sigma(\cdot)$ is the GeLU activation function. 


Finally, we use a Multi-Layer Perceptron (MLP) mapping the action probability tensor $\boldsymbol{P}_{a_t}$ to a unique result $a_t^\prime$ in the closed action set $\mathcal{A}$ as the action that a vehicle needs to perform in the current step:

\begin{equation}
\label{eq:7}
a_t^\prime = [{a^{(1)}_t}^\prime, {a^{(2)}_t}^\prime] = \text{MLP}_{\boldsymbol{W}_{a}}(\boldsymbol{P}_{a_t}),
\end{equation}
where ${a^{(1)}_t}^\prime$ is the normalized distance from the current location, ${a^{(2)}_t}^\prime \in [0^\circ, 360^\circ]$ is the direction that a vehicle headed to, and both of ${a^{(1)}_t}^\prime$ and ${a^{(2)}_t}^\prime$ make up the unique action that controls this vehicle, $\boldsymbol{W}_{a}$ is the training parameters.

Note that the difference between $359^\circ$ and $1^\circ$ is only 2 degrees when measuring angles.
Most common loss functions (\eg MAE Loss and MSE Loss) cannot describe this phenomenon well.
To train our framework \name\ effectively, we proposed a novel loss function, named Geospatial Loss (GeoLoss), to minimize the geospatial difference between the predicted action and the ground truth for this task, and our training target is to minimize the GeoLoss that we defined below:

\begin{equation}
\label{eq:8}
\begin{aligned}
\min Loss&(a_t^\prime , a_t)= \min \left( |{a^{(2)}_t}^\prime - a^{(2)}_t|^2, \right. \\
& \left. (360^\circ - |{a^{(2)}_t}^\prime  - a^{(2)}_t|)^2 \right) + |{a^{(1)}_t}^\prime - a^{(1)}_t|^2 .
\end{aligned}
\end{equation}

\section{Experiments}

This section conducts extensive experiments using 2 real-world datasets to evaluate the effectiveness of \name.
We first introduce the experimental settings.
Next, we compare \name\ with representative baselines.
Finally, the ablation study and a case study are introduced.

\subsection{Experimental Settings}

\noindent \textbf{Dataset.}
\quad
We use two datasets with different scales for experiments:
one is located in lower and midtown Manhattan, New York City, USA\footnote{https://data.cityofnewyork.us/Transportation/2018-Yellow-Taxi-Trip-Data/t29m-gskq/about\_data}, and the other larger dataset is the taxi trajectory data from the core area of Hangzhou, Zhejiang Province, CHN\footnote{Prviate dataset. To protect data copyright, we will share the full dataset through academic collaboration only.}. More details can be found in Appendix A.

\noindent \textbf{Metrics.}
\quad
We use the Euclidean distance metric, \emph{Error}, to assess the discrepancy between predicted actions and the driver’s actual driving intentions. Additionally, the \emph{empty-loaded rate} metric is employed to measure the efficiency of the car-hailing service, which is another widely used metric in transportation systems \cite{cao2021optimization}.

\noindent \textbf{Baselines.}
\quad
We compare \name\ with baselines from two different categories: (1) Online RL methods: 
MT \cite{robbennolt2023maximum} and
FTPEDEL \cite{wagenmaker2023leveraging};
(2) Offline RL methods: 
CQL \cite{kumar2020conservative},
TD3+BC \cite{fujimoto2021minimalist},
Decistion Transformer (DT) \cite{chen2021decision},
RLPD \cite{ball2023efficient},
latent offline RL \cite{hong2024learning}, and
SS-DT \cite{zheng2023semi};
(3) traditional vehicle dispatching systems: DGS \cite{DBLP:conf/atal/ChengJR18} and A-RTRS \cite{DBLP:conf/ijcai/RileyHY20}.
More details about these methods can be found in Appendix C.

\subsection{Implementation Detail}

Firstly, it's important to note that the correlation and influence between vehicles decrease exponentially with distance \cite{zhao2024asynchronous}.
To avoid network congestion, we only allow V2V communications between vehicles in adjacent regions. In addition, we limit the waiting time for vehicles to broadcast and receive V2V multi-hop messages across different regions to 1 second, ignoring any timed-out transmissions.
Typically, the delay for 1-hop 5G V2V data transmission ranges from 50ms to 200ms \cite{hakak2023autonomous, boualouache2023survey}, enabling at least 5-hop communication within this limitation. As the macro-level graph shown in Figure \ref{fig:3}, 5 communication hops allow vehicles to gather traffic information within about a 55km radius (city-level), which is sufficient for vehicle dispatching tasks.

Additionally, our experiments are simulated on the open-source software SUMO for secondary development and are deployed in a Linux server with two Intel(R) Xeon(R) Gold 6248R CPUs, 8 NVIDIA V100 32G graphics cards, and 800G memory. More details can be found in Appendix D.

\subsection{Overall Performance}

The performance of all the baselines in both two datasets is shown in
Table \ref{tab:baselines}, in terms of the two metrics we introduced before, \ie \emph{Error} and \emph{empty-loaded rate}.
We use \textbf{M} to present the Manhattan dataset and use \textbf{H} to stand for the Hangzhou dataset.
The performance of all methods is the average result of the last 100
epochs in a total of 1000 runs.

We can see that \name\ significantly reduces the error compared to other baselines, primarily due to our adoption of a more effective loss function for guiding the model during back-propagation and training. Unlike online reinforcement learning methods, nearly all offline RL approaches, including ours, outperform the online method by better utilizing offline data for effective training. Additionally, conventional offline RL methods (\eg TD3+BC, CQL, and RLPD) perform poorly on the larger Hangzhou dataset due to their inability to gather comprehensive traffic information within acceptable transmission delays. Although DT, latent offline RL, and SS-DT use similar stacked transformer decoder layers as our framework, they do not model driving behavior in scheduling tasks, limiting their accuracy.

\begin{table}[htb!]
\centering
\caption{Experimental results of different baselines.}
\label{tab:baselines}
\renewcommand\tabcolsep{2pt}
\begin{tabular}{ccccc}
\hline
& \multicolumn{2}{c}{Error (km)}    & \multicolumn{2}{c}{Empty-loaded Rate (\%)} \\
& \bfseries{M} & \bfseries{H} & \bfseries{M} & \bfseries{H} \\ \hline
MT & 0.3517 & 0.3929 & 38.23 & 47.52 \\
FTPEDEL & 0.3307 & 0.3451 & 36.03 & 42.74 \\ \hline
TD3+BC & 0.3371 & 0.3243 & 37.22 & 50.13 \\
CQL & 0.3125 & 0.2368 & 35.17 & 46.87 \\ 
DT & 0.3051 & 0.2573 & 33.49 & 41.45 \\ 
RLPD & 0.3213 & 0.3004 & 36.24 & 48.21 \\ 
Latent offline RL & 0.3117 & 0.2086 & 34.37 & 45.22 \\  
SS-DT & 0.3048 & 0.1843 & \textbf{32.25}* & 40.99 \\ \hline
DGS & 0.4125 & 0.1982 & 32.57 & 41.23 \\ 
A-RTRS & 0.3567 & 0.1957 & 32.39 & 41.04 \\
\hline
\textbf{\name} & \textbf{0.3044}* & \textbf{0.1582}* & 32.38 & \textbf{40.71}* \\ \hline
\end{tabular}

\small{
“*” indicates the statistically significant improvements (\ie two-sided t-test with $p < 0.05$) over the best baseline.

For all metrics: the lower, the better.
}

\end{table}

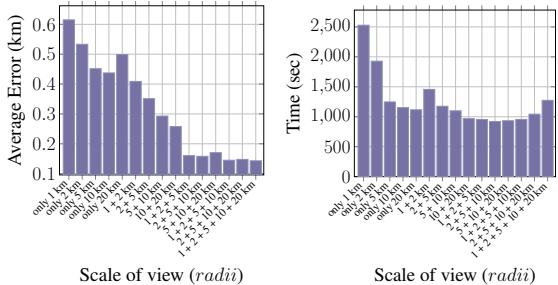
\begin{figure}[hbt!]
\centering
\subfigure[Error under 1-second data transmission limitation]{
\label{fig:multiview_a}
\resizebox{0.42\linewidth}{!}{
\begin{tikzpicture}
\begin{axis}[
ybar,
grid=major,
xmin = 0.3,
xmax = 15.6,
xtick = {1,2,3,4,5,6,7,8,9,10,11,12,13,14,15},
xticklabels={only 1 km,only 2 km, only 5 km, only 10 km, only 20 km, 1 + 2 km, 2 + 5 km, 5 + 10 km, 10 + 20 km, 1 + 2 + 5 km, 2 + 5 + 10 km, 5 + 10 + 20 km, 1 + 2 + 5 + 10 km, 2 + 5 + 10 + 20 km, 1 + 2 + 5 + 10 + 20 km},
xticklabel style={
    anchor=east,
    rotate=45,
    font=\normalsize
},
xlabel=Scale of view ($diameter$),
ylabel=Average Error (km),
font=\huge
]
\addplot [mycolorblue,fill=mycolorpurple] table [x=id,y=error,,col sep=comma] {data/multiview_error.csv};
\end{axis}
\end{tikzpicture}
}}
\subfigure[Overall training time for achieving sub-0.7 km error]{
\label{fig:multiview_b}
\resizebox{0.44\linewidth}{!}{
\begin{tikzpicture}
\begin{axis}[
ybar,
grid=major,
xmin = 0.3,
xmax = 15.6,
ymin = 0,
xtick = {1,2,3,4,5,6,7,8,9,10,11,12,13,14,15},
xticklabels={only 1 km, only 2 km, only 5 km, only 10 km, only 20 km, 1 + 2 km, 2 + 5 km, 5 + 10 km, 10 + 20 km, 1 + 2 + 5 km, 2 + 5 + 10 km, 5 + 10 + 20 km, 1 + 2 + 5 + 10 km, 2 + 5 + 10 + 20 km, 1 + 2 + 5 + 10 + 20 km},
xticklabel style={
    anchor=east,
    rotate=45,
    font=\normalsize
},
xlabel=Scale of view ($diameter$),
ylabel=Time (sec),
font=\huge
]
\addplot [mycolorblue,fill=mycolorpurple] table [x=id,y=num,,col sep=comma] {data/multiview_error.csv};
\end{axis}
\end{tikzpicture}
}
}
\caption{Different combinations of multiview graphs.}
\label{fig:multiview}
\vspace{-6mm}
\end{figure}

When comparing the metric of empty-loaded rate in Table \ref{tab:baselines}, our method ranks among the best ones.
However, our model needs to weigh the driver's personal driving behavior habits.
Therefore, an area with a slightly longer route that is more familiar to the driver has more chance of being selected for vehicle dispatching.
This caused the empty-loaded rate of \name\ to be slightly higher than the SS-DT method on the Manhattan dataset.
However, when the scale of the offline dataset becomes larger (Hangzhou dataset), our method has a stronger ability to find the optimal dispatching strategy and achieve the best performance while satisfying the driver's driving behavior.

\subsection{Ablation Study}

\noindent \textbf{The effectiveness of multiview graph.}
\quad
To better understand the role of multiview graph learning in \name, we divide the road network into regions with diameters of 1 km and 2 km (micro-level), 5 km (meso-level), and 10 km and 20 km (macro-level). We then extract 5 graphs with varying traffic features based on these granularities. Under the same 1-second V2V data transmission delay previously mentioned, we sequentially use various combinations of these graphs as inputs to conduct experiments. The results are presented in Figure \ref{fig:multiview_a}.

The results indicate that the error in graph learning using a single view is significantly higher than that of multiview graph learning methods. Moreover, when comparing different scales, it is evident that model performance improves as the granularity of the scale decreases.

To further explore the relationship between computational latency (including V2V communication and model training time) and multiview graphs, we compared the training time of each model when achieving a scheduling error of 0.7 km, as shown in Figure \ref{fig:multiview_b}.
It also verifies that the multiview graph-based learning method could be more efficient than the single-view graph-based learning method.

In addition, by analyzing the average error in Figure \ref{fig:multiview_a} alongside the time cost in Figure \ref{fig:multiview_b}, we selected three multiview graphs with diameters of 2 + 5 + 10 km to model the traffic of road networks efficiently.


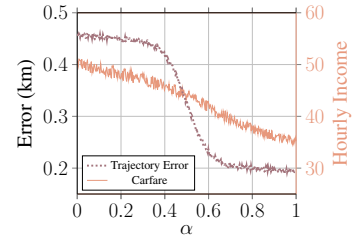
\begin{figure}[h]
\centering
\resizebox{0.55\linewidth}{!}{
\begin{tikzpicture}
\color{mycolororange}
\begin{axis}[
smooth,
grid=major,
axis y line*=right,
xmin=0,
xmax=1,
xtick = {0,0.2,0.4,0.6,0.8,1.0},
xticklabels={,,,,,},
tick align=outside,
ylabel=Hourly Income,
ylabel style ={mycolororange},
ymin=25,
ymax=60,
font=\huge
]
    \addplot [
    mark=none,
    thick,
    mycolororange] table [x=alpha, y=fare,, col sep=comma] {data/alpha.csv};
    \label{ax_fare}
\end{axis}
\color{black}
\begin{axis}[
smooth,
axis y line*=left,
tick align=outside,
ylabel=Error (km),
ymin=0.15,
ymax=0.50,
xmin=0,
xmax=1,
xtick = {0,0.2,0.4,0.6,0.8,1.0},
xlabel=$\alpha$,
font=\huge,
transpose legend,
legend columns=1,
legend style={at={(0.28,0.02)},anchor=south,font=\normalsize, text opacity=1, fill opacity=0.5}
]
    \addplot [
    mark=none,
    ultra thick,dotted,
    mycolorbrown] table [x=alpha, y=error,, col sep=comma] {data/alpha.csv};
\addlegendentry{Trajectory Error}
\addlegendimage{/pgfplots/refstyle=ax_fare}\addlegendentry{Carfare}
\end{axis}
\end{tikzpicture}
}
\caption{Results under different settings of $\alpha$.}
\label{fig:alpha}
\end{figure}

\noindent \textbf{The Influence of Driving Behavior.}
\quad
To verify the influence of driving behavior, we conduct
experiments on this method alone via setting different hyperparameter $\alpha$ defined in Equation \eqref{eq:5}.
The experimental result is shown in Figure \ref{fig:alpha}.
When the weight of driving behavior increases, the error of the predicted vehicle trajectory decreases.
At this time, the vehicle follows the path given by the offline data and cannot explore the path that can generate higher income in accordance with the vehicle's driving behavior.
Conversely, when $\alpha$ is close to 0, many passenger loading locations seriously deviate from the roads and regions familiar to the driver although the vehicle can receive more orders.
These potential issues are more likely to cause traffic accidents.
From Figure \ref{fig:alpha} we can see that the trajectory error significantly drops when $\alpha$ is between 0.4-0.7.
Therefore, in this paper, we set $\alpha=0.67$ to let the dispatch strategy increase the driver's income as much as possible while satisfying each driver's driving behavior.

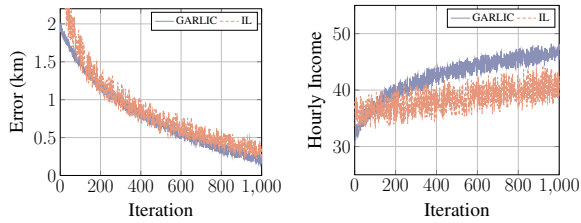
\begin{figure}[htb!]
\centering
\subfigure[Training error with epoch increases]{
\label{fig:add_cmp_a}
\resizebox{0.45\linewidth}{!}{
\begin{tikzpicture}
\begin{axis}[
smooth,
grid=major,
xlabel=Iteration,
ylabel=Error (km),
ymin=0,
ymax=2.2,
xmin=0,
xmax=1000,
legend columns=2,
legend style={font=\normalsize, text opacity=1, fill opacity=0.5},
font=\huge
]
    \addplot [mark=none,mycolorblue] table [x=id, y=model,, col sep=comma] {data/add_cmp.csv};
    \addplot [mark=none,mycolororange, thick, densely dashed] table [x=id, y=cmp,, col sep=comma] {data/add_cmp.csv};
    \legend{\name, IL}
\end{axis}
\end{tikzpicture}
}}
\subfigure[Expected hourly income with epoch increases]{
\label{fig:add_cmp_b}
\resizebox{0.45\linewidth}{!}{
\begin{tikzpicture}
\begin{axis}[
smooth,
grid=major,
xlabel=Iteration,
ylabel=Hourly Income,
ymin=25,
ymax=55,
xmin=0,
xmax=1000,
legend columns=2,
legend style={font=\normalsize, text opacity=1, fill opacity=0.5},
font=\huge
]
    \addplot [mark=none,mycolorblue] table [x=id, y=model,, col sep=comma] {data/add_cmp2.csv};
    \addplot [mark=none,mycolororange, thick, densely dashed] table [x=id, y=cmp,, col sep=comma] {data/add_cmp2.csv};
    \legend{\name, IL}
\end{axis}
\end{tikzpicture}
}
}
\caption{Comparison between \name\ and IL.}
\label{fig:add_cmp}
\end{figure}

\noindent \textbf{The efficiency of dynamic rewards.}
\quad
In this paper, we propose a dynamic reward function to score different vehicles' driving behavior and expected carfare.
This reward function is also the core difference of \name\ from vehicle dispatching-based Imitation Learning (IL) methods \cite{li2022driver}.
In IL methods, a human expert is always needed to provide high-quality offline data instead of the reward function.
\revise{To verify the efficiency of rewards, we conduct experiments on \name\ and an IL method that only takes traffic states as input and outputs the predicted actions.}


The experimental results are shown in Figure \ref{fig:add_cmp}.
The results show that our method is more robust and has lower trajectory error than the IL method when the number of training iterations is greater than 800.
Furthermore, \name\ is able to efficiently explore trajectories with higher incomes, while IL only learns traffic characteristics from the provided data.

\subsection{Case Study}

\begin{figure}[!t]
\centering
\setlength{\belowcaptionskip}{0pt}
\includegraphics[width=0.85\linewidth]{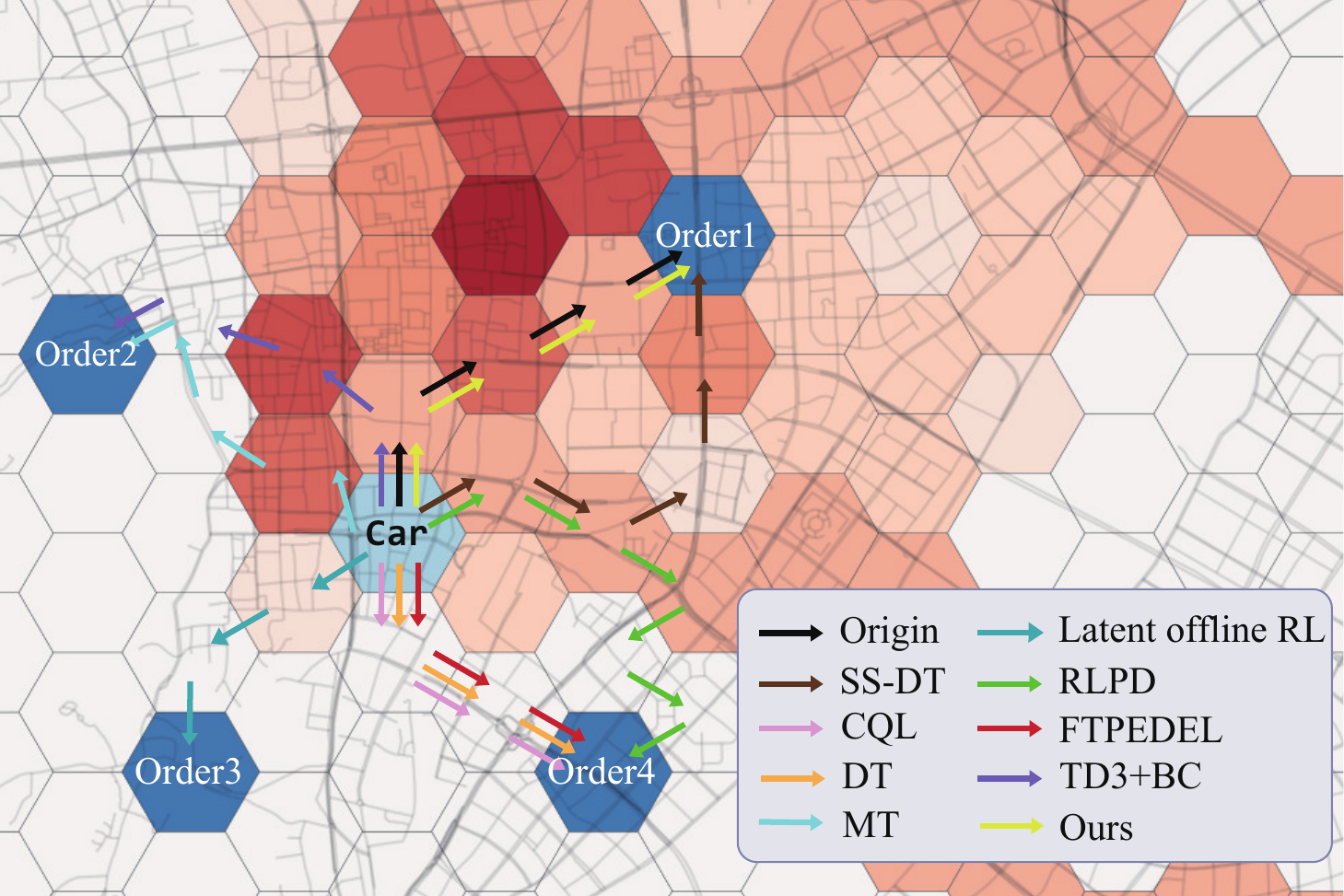}
\caption{An example of vehicle dispatching.}
\label{fig:case}
\vspace{-6mm}
\end{figure}

We randomly choose an online car-hailing car in Hangzhou as our experimental object.
We simulated the vehicle dispatching routes using different methods to compare with the ground truth (Origin), as shown in Figure \ref{fig:case}.
We selected a local area in Hangzhou for visualization.
Different shades of red in honeycomb grids indicate the length of time the vehicle stayed in one week of history.
When the area has no color, the vehicle has not been to this area that week.
It represents the driver's personalized driving behavior.
The vehicle is currently located in the bright blue grid, and we visualize 4 dark blue areas with ride-hailing demand in the next 15 minutes.
We use arrows of different colors to indicate the calculation results of different models.

As can be seen from Figure \ref{fig:case}, Order 1 is farther from the departure point of the vehicle compared to Orders 2, 3, and 4.
Since Order 4 is in the city center, most methods select this area as the vehicle pick-up point. However, there is a direct arterial road between Order 2 and the vehicle's departure location, so some methods choose to dispatch the vehicle to where Order 2 is located.
Our method analyzes the driver's driving behavior and finds that the most suitable place to pick up passengers is the area where order 1 is located. Meanwhile, only the results calculated by our method are consistent with the actual vehicle trajectory, which shows the effectiveness of our proposed method.

\section{Conclusion}

In this paper, a novel framework called \name\ is proposed to address the problem of vehicle dispatching while considering the driving behavior of drivers at the same time.
Specifically, it can be divided into three modules, \ie the hierarchical traffic state representation module for traffic state extraction, the dynamic reward generation module for driving behavior as well as carfare analysis, and the GPT-augmented dispatching policy learning module for balancing vehicle supply and passenger demand.
The model achieves a response in seconds under multiple real datasets and has excellent performance.
In the future, we hope to combine the Kafuka engine and cloud-edge collaboration technologies to further optimize the information transmission of each node in vehicle dispatching, achieve a quick response of hundreds of milliseconds, and improve the driver's order acceptance and user's riding experience.

\section{Acknowledgments}

This research was partially supported by Huawei (Huawei Innovation Research Program), Research Impact Fund (No.R1015-23), APRC - CityU New Research Initiatives (No.9610565, Start-up Grant for New Faculty of CityU), CityU - HKIDS Early Career Research Grant (No.9360163), Hong Kong ITC Innovation and Technology Fund Midstream Research Programme for Universities Project (No.ITS/034/22MS), Hong Kong Environmental and Conservation Fund (No. 88/2022), and SIRG - CityU Strategic Interdisciplinary Research Grant (No.7020046), Huawei (Huawei Innovation Research Program).

\newpage

\section{A\quad Dataset}
\label{app:b}

\begin{figure}[ht]
\centering
\subfigure[Manhattan]{
\includegraphics[width=0.45\linewidth]{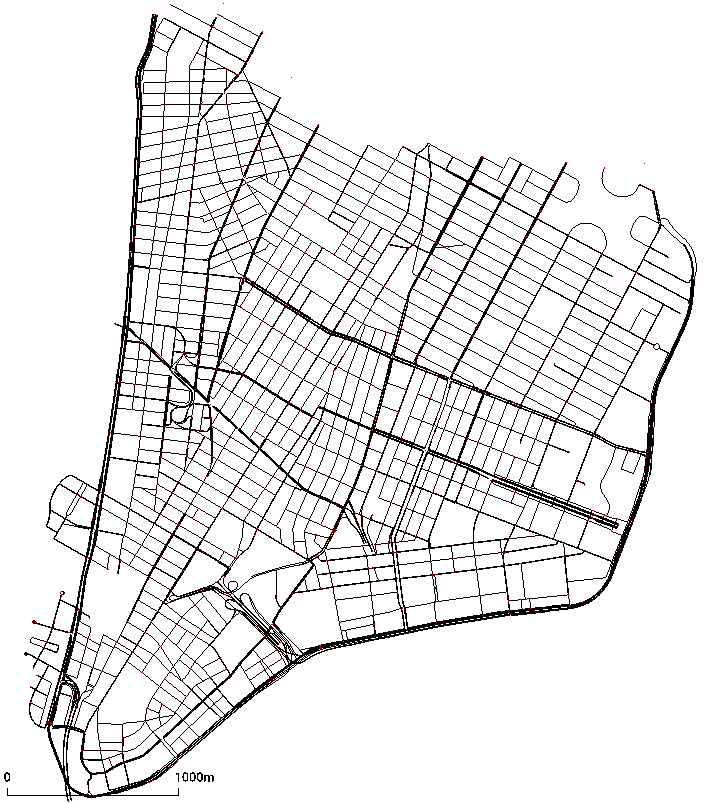}
\label{fig:rdmap_a}
}
\subfigure[Hangzhou]{
\includegraphics[width=0.45\linewidth]{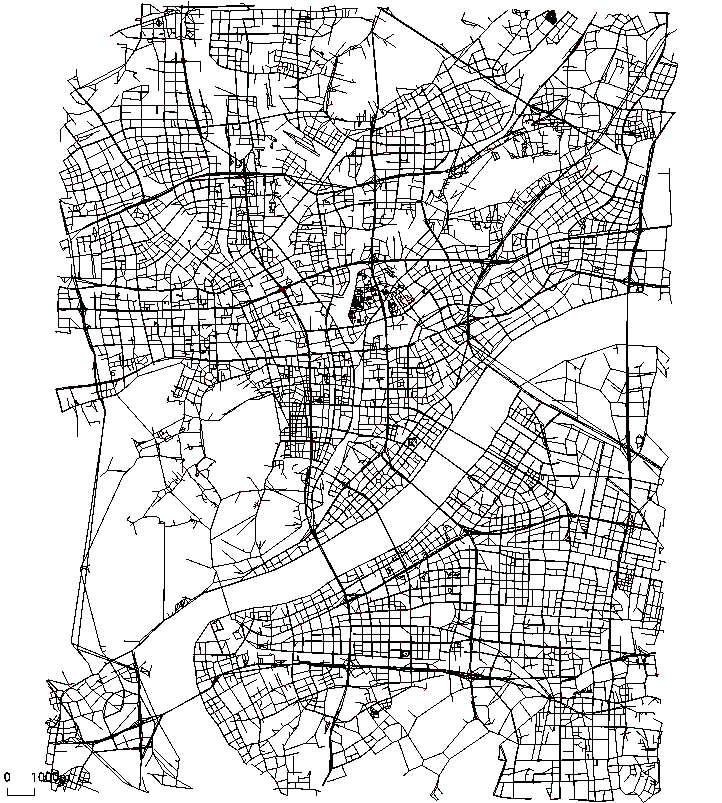}
\label{fig:rdmap_b}
}
\caption{Road networks of different datasets.}
\label{fig:rdmap}
\end{figure}

We use two datasets located in Manhattan\footnote{https://data.cityofnewyork.us/Transportation/2018-Yellow-Taxi-Trip-Data/t29m-gskq/about\_data} and Hangzhou\footnote{Prviate dataset. To protect data copyright, we will share the full dataset through academic collaboration only.}, and the road networks of these two cities are shown in Figure \ref{fig:rdmap}.
The statistics of datasets we used in this paper are listed in Table \ref{tab:1}. In the first dataset, we collect trajectories from 350 taxis within 4 hours and all car-hailing orders in this period that indicate the start and end locations.
As for the second dataset, we collect trajectories from 9041 taxis within 1 month and also all car-hailing orders in this period.
For all datasets, we partition the training, validation, and test sets in a 6:3:1 ratio based on chronological order.

\begin{table}[htb!]
\centering
\caption{Statistics of datasets.}
\label{tab:1}
\begin{filecontents*}{data/dataset.csv}
title, Manhattan, Hangzhou
Number of taxis, 350, 9041
Coverage ($\mathrm{km}^2$), 18, 900
Number of sub-areas, 19, 928
Trajectory records, $84\,000$, $781\,142\,400$
Order records, $6\,317$, $15\,144\,840$
Data Period, 4 Hours, 30 Days
\end{filecontents*}
\csvreader[
   tabular=ccccccc,
   table head=\hline
  
  
   \multicolumn{1}{c}{} & \multicolumn{1}{c}{\bfseries{Manhattan (M)}} & \multicolumn{1}{c}{\bfseries{Hangzhou (H)}} \\
  
  
   \hline,
   late after line={\\},
   late after last line= \\
   \hline 
]{data/dataset.csv}{}{\csvlinetotablerow}
\end{table}

\section{B\quad Data Preprocessing}

In this part, we introduce data preprocessing in detail, which mainly contains three parts:
map matching, anomalous trajectory detection and filter, and trajectory segmentation.


First, a statistically based anomaly trajectory identification method is applied to improve trajectory data quality.
To be specific, we deploy map matching methods \cite{yu2022high,wu2020map} to initially map the GPS points to the road segments to solve the problem of GPS drift.
Next, we use the shortest path algorithm to calculate the actual trajectory between every two mapped points and add auxiliary GPS anchors at each intersection to ensure that the road segments of the trajectory are continuous.

Next, we calculate the average speed of a vehicle traveling on a trajectory composed of time series GPS points
and then remove the GPS points that cause the average vehicle speed to be 20\% higher than the road speed limit.
When modeling the driver's driving behavior, we only extract the trajectory data during the idling period of the vehicle for pre-training.
When training the dynamic dispatching model, we use the complete trajectory data since the carfare and driving behavior jointly constitute the dynamic reward.

As mentioned in the third section, 
when we train the trajectory data in the form of sequence data, the cumulative error generated during the training process will reduce the model's convergence speed.
This effect will increase exponentially with increasing sequence data step size.
Taking a vehicle to generate a GPS track point every 30 seconds as an example, the trajectory length per month can reach 86400, which is a huge number.
In order to reduce the impact of ultra-long sequence data on the model convergence speed, we adopt the method of position embedding coding and trajectory segmentation training: we define the maximum sequence data length $Leng$ ($Leng=1024$ is used in this paper).
Then we randomly select multiple trajectory sequences (1000-10000 samples) with any random length less than or equal to $Leng$ from each vehicle trajectory as the model training sample set.

\section{C\quad Baseline Details}
\label{app:baselines}

The details of baseline models are described as follows:

\noindent \textbf{MT} \cite{robbennolt2023maximum}.
\quad
It uses statistical methods to build an optimization model of vehicle dispatching, which builds on the dynamic queuing model design of Kang and Levin which provides
a maximum stability dispatch policy for SAVs.

\noindent \textbf{FTPEDEL} \cite{wagenmaker2023leveraging}.
\quad
This method is formulated as an integer linear program that minimizes the total travel time of SAVs while also minimizing a penalty related to vehicle imbalance in the network.

\noindent\textbf{CQL} \cite{kumar2020conservative}.
\quad
It is an offline reinforcement learning method that addresses the limitations of overestimating values induced by the distributional shift between the dataset and the policy.

\noindent\textbf{TD3+BC} \cite{fujimoto2021minimalist}.
\quad
This method is an efficient offline reinforcement learning method.
It only adds a regular term of behavior cloning (BC) to the value function and normalizes the state to achieve the same accuracy as other complex offline reinforcement learning methods.

\noindent\textbf{Decistion Transformer (DT)} \cite{chen2021decision}).
\quad
This method is the pioneering work of RL Transformer.
It regards the RL problem as a sequence modeling problem, abandons the calculation modes such as Policy Gradient in the traditional RL method, and improves the calculation efficiency.

\noindent \textbf{RLPD} \cite{ball2023efficient}.
\quad
This method builds upon standard off-policy RL algorithm and incorporates symmetric sampling, layer normalization, sample-efficient learning, and large ensembles.

\noindent \textbf{Latent offline RL} \cite{hong2024learning}.
\quad
This method utilizes Conservative Q-Learning (CQL) with a low-dimensional representation learning to obtain human behavior, allowing the agent to adapt its strategy as the human behavior changes over time.

\noindent \textbf{SS-DT} \cite{zheng2023semi}.
\quad
It introduces a semi-supervised offline reinforcement learning approach that leverages both labeled and unlabeled data to train reinforcement learning agents, and we use DT as its base RL model.

\noindent \textbf{DGS} \cite{DBLP:conf/atal/ChengJR18}.
\quad
This method consists of four key modules: the Data Stream Handler, which cleans and aligns GPS data to ensure location accuracy; the Demand and Supply Prediction Engine, which uses a multi-level logistic regression model to estimate demand probabilities across different streets and time periods; the Multi-Agent Recommendation Engine, which generates personalized recommendations for drivers through an optimization model to balance city-wide demand and supply; and the Mobile Application, which delivers these recommendations to drivers, enhancing their navigation experience.

\noindent \textbf{A-RTRS} \cite{DBLP:conf/ijcai/RileyHY20}.
\quad
This paper presents an end-to-end framework for real-time scheduling in large-scale ride-sharing systems. The framework integrates optimization algorithms, machine learning, and model predictive control (MPC) to optimize vehicle scheduling, path planning, and idle vehicle relocation, with the goal of minimizing average waiting times.

\section{D\quad Implementation Details}
\label{app:imp}

We have three parts in the proposed framework, such as the multiview graph representation, dynamic reward modeling, and a GPT-augmented model. 
In the multiview graph representation, we set the hidden dimension of each view of the graph convolutional layer as 128.
Besides, $\alpha$ in the dynamic reward function is 0.65.
We use GPT2 as our foundation GPT model, and the maximum length of trajectory input is 1024 when training the entire data set.
The rate $dp = 0.5$ is set for all dropout layers in \name.
In addition, Adam is used as the optimizer for all models, and we use the default parameter for it, \emph{i.e.,} $\beta_1=0.9$, $\beta_2=0.999$, $\epsilon=10^{-8}$, $lr=10^{-3}$. 

Our experimental platform is based on the open-source software SUMO for secondary development and is deployed in a Linux server with two Intel(R) Xeon(R) Gold 6248R CPUs, eight NVIDIA TESLA V100 32G graphics cards, and 800G memory.
In addition, we incorporate a computing unit located inside each vehicle as an edge computing engine to assist our algorithms in performing real-time computations.
Noting that the Hangzhou dataset is extremely large, we deployed the distributed processing platform, Spark, in an additional platform (120 executors with 30GB memory and 80 CPUs per node) to simulate the real-time operation of all vehicles and submit traffic information once a time to the above-mentioned server for training/inference through network requests.

\section{E\quad Training \& Testing Process}
\label{subsec:E}

\subsection{Training Process}

In this subsection, we introduce the training process of the \name. 
As shown in Algorithm \ref{alg:1}, we deploy a two-stage sequential learning framework since the GPT-based model usually has the limitation of slow convergence.
The training process can be divided into two parts, and every agent shares the same model parameters: 
traffic state representation \& dynamic reward learning, and a GPT-augmented model for policy learning.

The first part extracts sequences of states, actions, and rewards for each vehicle agent, shown in lines 1-10.
The second part is used to learn the relationship between the action and the environment state, as well as the reward.
It aims to predict the expected action under the environmental state at the current moment, shown in lines 11-19.

\begin{algorithm}[htp!]
\caption{\name\ Training Process}
\label{alg:1}
\begin{algorithmic}[1]
    \REQUIRE 
        Multiview Graph and its features $\mathcal{G}_i$ \& $\boldsymbol{X}_{i,t}$, vehicle trajectory $\tau$ and car-harling fare $x_{\text{fare},t}$, $t \in [T]$.
    \ENSURE all parameters $\boldsymbol{W}$ in the model. \\

    \textbf{Traffic state representation \& Dynamic reward learning:} \\
    
    \STATE Initialize parameters $\boldsymbol{W}_i$, $\boldsymbol{W}_{\text{GRU}}$
    
    \FOR{Each timestep $t=1, \cdots, T$}

    \STATE Calculate multi-view graph representation: $\boldsymbol{Emb}_{\mathcal{G}_i, t}$ \\ where $i = \{$ micro, meso, macro $\}$

    \STATE Calculate location embedding for each vehicle: $\boldsymbol{Emb}^u_{\text{loc},t}$

    \STATE Calculate the traffic state around each vehicle: $s^u_t$

    \STATE Calculate the probability that the location belong to a given vehicle: $p_t^u$

    \STATE Update parameter: $\boldsymbol{W}_i$, $\boldsymbol{W}_{\text{GRU}}$

    \STATE Calculate the dynamic reward: $r_t^u$

    \STATE Calculate the action compared to the location of the vehicle at previous timestep: $a^u_t$

    \ENDFOR
    
    \textbf{GPT-augmented model training:} \\

    \STATE Initialize parameters $\boldsymbol{W}_{\text{fare}}$, $\boldsymbol{W}_{x}$, $\boldsymbol{W}_{y}$, $\boldsymbol{W}_{z}$, $\boldsymbol{W}_{a}$

    \FOR{Each training iteration}

    \FOR{Each vehicle $u$}
    
    \STATE Calculate: $\boldsymbol{Emb}_r = [\boldsymbol{Emb}_{r_1}, \cdots, \boldsymbol{Emb}_{r_T}]$

    \STATE Predict: $\boldsymbol{P}_{a} = [\boldsymbol{P}_{a_1}, \cdots, \boldsymbol{P}_{a_T}]$

    \STATE Action choose: $a^\prime = [a^\prime_1, \cdots, a^\prime_T]$

    \STATE Update parameters: $\boldsymbol{W}_{\text{fare}}$, $\boldsymbol{W}_{x}$, $\boldsymbol{W}_{y}$, $\boldsymbol{W}_{z}$, $\boldsymbol{W}_{a}$
    
    \ENDFOR

    \ENDFOR
    
    \STATE \textbf{return} all trained parameters $\boldsymbol{W}$
\end{algorithmic}
\end{algorithm}

\subsection{Testing Process}

As an extension to the section D, this section continues to introduce the testing process of the framework.
During the testing period,
we have all trained parameters $\boldsymbol{W}$.
Then for a vehicle $u$, we apply the pre-training module to calculate traffic state $s_t^u$ and expected dynamic reward $a_t^u$ by giving the real-time observation $X_t^u$ and trajectory $\tau^u$.
Next, the GPT-augmented model is deployed to predict the possible dispatching action $a_t^u$ as shown in Algorithm \ref{alg_test}.

In addition, we implement a greedy algorithm \cite{zheng2019auction} to assign passenger orders to vehicles.
When there are passengers waiting to be picked up on the side of the road, we will sequentially select the vehicle closest to the passenger and whose current vehicle status is empty as the order-receiving vehicle.
This ensures a relatively short time for passengers waiting at the roadside.

\begin{algorithm}
\caption{\name\ Testing Process}
\label{alg_test}
\begin{algorithmic}[1]
\REQUIRE
    Trained global parameter $\boldsymbol{W}$,
    history vehicle trajectory $\tau^{(\text{his})}$,
    history carfare $x_{\text{fare}}^{(\text{his})}$,
    history multiview graphs $\mathcal{G}^{i,(\text{his})}$,
    history traffic features $\boldsymbol{X}^{(i,\text{his})}$.
        
\ENSURE Real-time dispatching action $a^u_t$ for vehicle $u$. \\

\textbf{Each vehicle $u$:}
\STATE Observe the real-time traffic feature: $\boldsymbol{X}_{t}^{i}$

\STATE Update history traffic features $\boldsymbol{X}^{i,(\text{his})}$

\STATE Calculate traffic states $s_t$

\STATE Generate expected dynamic reward $r_t$

\STATE Sample an action $a_t$ by GPT-augmented model

\STATE \textbf{return} $a^u_t$ to vehicle $u$ for relocation

\end{algorithmic}
\end{algorithm}

\section{F\quad Additional Ablation Study}

\begin{figure}[htb!]
\centering
\resizebox{0.6\linewidth}{!}{
\begin{tikzpicture}
\begin{axis}[
smooth,
grid=major,
xlabel=Iteration,
ylabel=Error (km),
ymin=0,
xmin=1,
xmax=1000,
legend columns=2,
legend style={font=\normalsize,at={(0.54,0.55)},anchor=south, text opacity=1, fill opacity=0.5},
font=\huge
]
    \addplot [mark=none,mycolor91] table [x=x, y=y1,, col sep=comma] {data/baseline_error.csv};
    \addplot [mark=none,mycolor92] table [x=x, y=y2,, col sep=comma] {data/baseline_error.csv};
    \addplot [mark=none,mycolor93] table [x=x, y=y3,, col sep=comma] {data/baseline_error.csv};
    \addplot [mark=none,mycolor94] table [x=x, y=y4,, col sep=comma] {data/baseline_error.csv};
    \addplot [mark=none,mycolor95] table [x=x, y=y5,, col sep=comma] {data/baseline_error.csv};
    \addplot [mark=none,mycolor96] table [x=x, y=y6,, col sep=comma] {data/baseline_error.csv};
    \addplot [mark=none,mycolor97] table [x=x, y=y7,, col sep=comma] {data/baseline_error.csv};
    \addplot [mark=none,mycolor98] table [x=x, y=y8,, col sep=comma] {data/baseline_error.csv};
    \addplot [mark=none,mycolor99] table [x=x, y=y9,, col sep=comma] {data/baseline_error.csv};
    \legend{CQL,TD3+BC,DT,FTPEDEL,MT,RLPD,Latent offline RL,SS-DT,\textbf{\name}}
\end{axis}
\end{tikzpicture}
}
\caption{Trends in error over training iteration.}
\label{fig:error_compare}
\vspace{-3mm}
\end{figure}
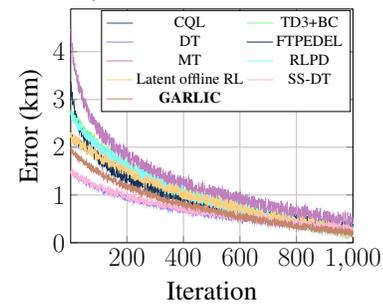

\noindent \textbf{Dispatching Performance.}
\quad
As mentioned in the previous study \cite{fujimoto2021minimalist}, the advantage of the offline reinforcement learning method is that its training time could be much shorter than that of the online reinforcement learning method, and it can make full use of valuable features in the training data.
To observe this phenomenon more intuitively, we visualize all methods' training errors with training epoch increases, as shown in Figure \ref{fig:error_compare}.
We can see that online RL-based methods, such as MT, have the largest errors at the beginning.
This is because these methods need to interact with the environment continuously to discover the optimal policy, and this process is very time-consuming.
Note that the convergence speed of FTPEDEL is significantly higher than that of CQL, TD3+BC, latent offline RL, and RLPD methods after several epochs.
This may be because FTPEDEL continuously explores a series of better strategies at that time.
Due to the disadvantage of the traditional offline reinforcement learning method, offline RL-based methods cannot interact with the environment, so heuristically learning the optimal strategy from a limited dataset is difficult.
Our method, DT, and SS-DT adopt the transformer structures to parallelize the training, dramatically reducing the model's initial training error.
From Figure \ref{fig:error_compare}, the dispatching error of DT and SS-DT is lower than our method within the first 600 epochs, which is mainly due to their model structure:
both of them take the reward, state, and action as the input of the model at the same time, while our method only takes the reward and state as the input.
Due to the addition of the real action at the previous moment as the input of the model,
those methods can significantly reduce the cumulative error of the model when learning time series data.
For trajectory data with the same sequence length, the token length of either DT or SS-DT as input will be 1.5 times that of our method.
Our method could effectively train more extended trajectory data than DT and SS-DT when using models with similar parameter scales for simultaneous training.

\begin{figure}[ht]
\centering
\resizebox{0.6\linewidth}{!}{
\begin{tikzpicture}
\begin{axis}[
smooth,
grid=major,
xlabel=Number of cars : Number of orders,
ylabel=Rate (\%),
ymax=100,
ymin=0,
xmin=-10,
xmax=10,
xtick = {-10,-8,-6,-4,-2,0,2,4,6,8,10},
xticklabels={1:10,1:8,1:6,1:4,1:2,1:1,2:1,4:1,6:1,8:1,10:1},
xticklabel style={
    anchor=north,
    font=\large 
},
legend columns=1,
legend style={font=\normalsize,at={(0.65,0.02)},anchor=south, text opacity=1, fill opacity=0.5},
font=\huge
]
\addplot [mycolorblue,mark=*] table [x=x, y=load,, col sep=comma] {data/rate.csv};
\addplot [mycolororange, mark=*] table [x=x, y=order,, col sep=comma] {data/rate.csv};
\legend{Empty-loaded Rate, Order Acceptance Rate}
\end{axis}
\end{tikzpicture}
}
\caption{Results under different vehicle-order ratios.}
\label{fig:ratio}
\end{figure}

\begin{figure*}[htb!]
\centering
\subfigure[Before deployment: The darker the grid, the greater imbalance between supply and demand is observed.]{
\includegraphics[width=0.88\linewidth]{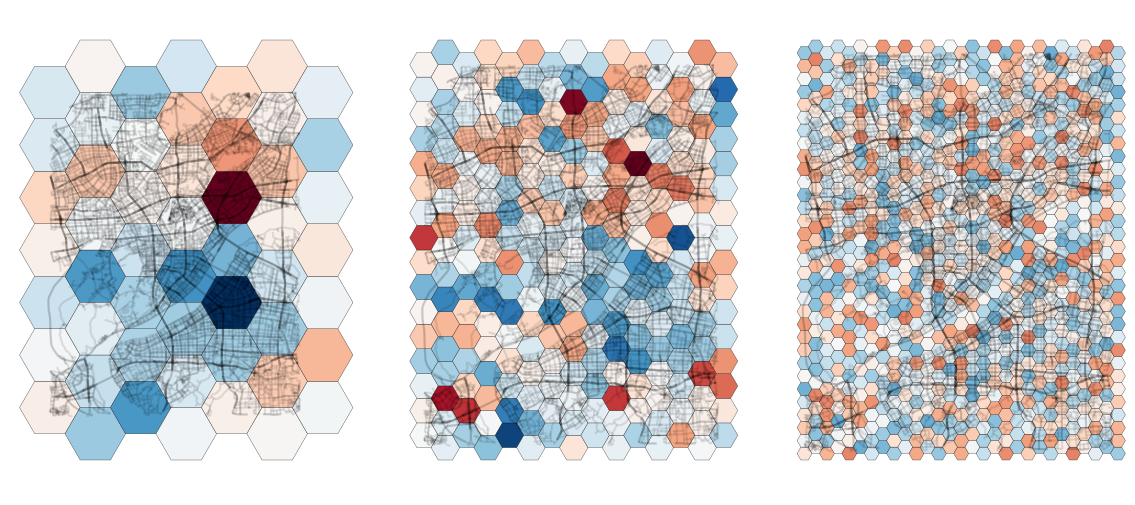}
\label{fig:deployment_a}
}
\subfigure[After deployment: The imbalance between supply and demand has been greatly alleviated.]{
\includegraphics[width=0.88\linewidth]{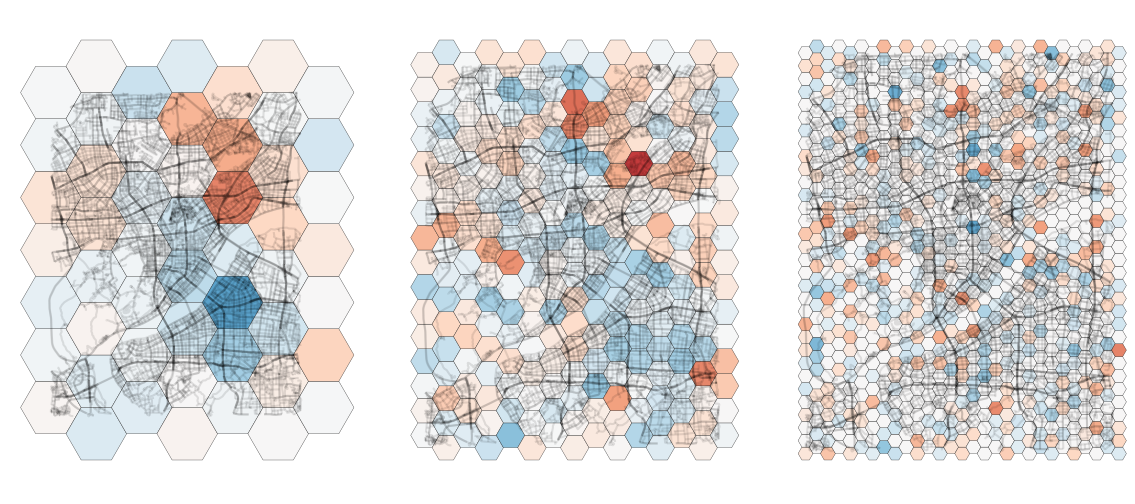}
\label{fig:deployment_b}
}
\caption{Result of vehicle dispatching.}
\label{fig:deployment}
\end{figure*}

\noindent \textbf{Ratio of car number and order number.}
\quad
In practice, there is also a close relationship between the number of vehicles operating in the road network and the number of demands for orders in real-time.
It is easy to imagine that when there are far more vehicles in the road network than the order demand, most vehicles will not be able to receive passengers due to too few orders.
On the other hand, when the number of vehicles in the road network is far less than the order demand, passengers in some remote areas may be unable to get a car-hailing service.
In both extreme cases, the vehicle dispatching algorithm cannot play its role fully.
To find the most suitable settings for deploying the model, we select different ratios of cars and orders to conduct experiments in the same simulation environment, as shown in Figure \ref{fig:ratio}.

\begin{table}[htb!]
\caption{Notations in this paper.}
\label{tab:notation}
\renewcommand\tabcolsep{5pt}
\begin{filecontents*}{data/notation.txt}
Notation; Description
$\tau$;The set of vehicle trajectory
$\mathcal{G}$;Graph of the road network
$\boldsymbol{V}$;The node set, $\boldsymbol{V} = \{ v_1, v_2, \cdots \}$
$\boldsymbol{E}$;The edge set, $\boldsymbol{E} = \{ e_1, e_2, \cdots \}$
$u$; A specific vehicle 
$\boldsymbol{X}$;The traffic features
$t$; The decision timestep in RL
$\mathcal{A}$; The set of actions, $\mathcal{A} = \{a_0, a_1, \cdots\}$
$\mathcal{S}$; The set of traffic states, $\mathcal{S}= \{ s_0, s_1, \cdots \}$
$\mathcal{R}$; The set of rewards, $\mathcal{R} = \{ r_1, r_2, \cdots \}$
$\mathcal{\pi}$; The vehicle dispatch policy
$\boldsymbol{Emb}$; The embedding matrix
$GeoHash(\cdot)$; The GeoHash algorithm
$\mathrm{GCN}(\cdot))$; Graph convolutional networks
$\mathrm{Decoder}(\cdot)$; The transformer decoder block
$\mathrm{MLP}(\cdot)$; The multi-layer perception model
$\boldsymbol{W}$; The learnable model parameters
$\sigma(\cdot)$; The activation function
$Loss$; The loss function
\end{filecontents*}
\csvreader[
separator=semicolon,
  tabular=cc,
  table head=\hline
 \multicolumn{1}{c}{\bfseries{Notation}} & \multicolumn{1}{c}{\bfseries{Description}} \\
  \hline,
  late after last line= \\
  \hline 
]{data/notation.txt}{}{\csvlinetotablerow}
\end{table}

In this paper, the order acceptance rate indicates the ratio of the number of orders that a vehicle picks up passengers within 10 minutes to the total number of real-time orders,
and the empty-loaded ratio refers to the total idle time of all vehicles in the road network as a percentage of the sum of each vehicle's total running time.
We can see that when the ratio of the number of vehicles to the number of orders is between 2:1 and 4:1, our method can achieve a lower empty-loaded rate while ensuring the timely receipt of passengers.

\section{G\quad Visualization}

As an extension of the case study, we visualize the supply and demand of the multi-view road network before/after our algorithm is deployed in this section, as shown in Figure \ref{fig:deployment}.

In Figure \ref{fig:deployment}, we mark the region as blue when the number of vehicles in the area exceeds the real-time order demand in the current region, otherwise, we mark it as red.
The darker the color of a region, the more unbalanced the supply and demand in this region, and only when supply is precisely equal to demand, the region will be transparent.
From Figure \ref{fig:deployment_a}, we can see that dividing the road network view with a larger granularity can show the supply and demand relationship of a city in the geographical distribution at a more macro level, even when the supply and demand relationship in the fine-grained road network view seems to be irregular.
However, the fine-grained road network view can obviously provide more features describing the distribution of supply and demand within the city, and these features are also critical to the feature extraction and prediction of \name.

Comparing Figure \ref{fig:deployment_b} to Figure \ref{fig:deployment_a}, we can see that there is a significant reduction (colored areas become lighter) in the phenomenon of supply-demand imbalance regions, which shows the effectiveness of our model.

\section{H\quad Notations}

In this section, we summarize all notations in the main paper and list them in Table \ref{tab:notation}.

\end{document}